# Image-based Face Detection and Recognition:

# "State of the Art"

Faizan Ahmad[1], Aaima Najam[2] and Zeeshan Ahmed[3]

[1] Department of Computer Science & Engineering, Beijing University of Aeronautics & Astronautics
Beijing, 100000, China
**faizan.ahmad.1988@gmail.com**

[2] Department of Computer Science, COMSATS Institute of Information Technology
Lahore, 54000, Pakistan
**aaimanajam@yahoo.com**

[3] Department of Information Technology, Education University
Lahore, 54000, Pakistan
**zeeshanahmeed@gmail.com**

**Abstract**
Face recognition from image or video is a popular topic in biometrics research. Many public places usually have surveillance cameras for video capture and these cameras have their significant value for security purpose. It is widely acknowledged that the face recognition have played an important role in surveillance system as it doesn't need the object's cooperation. The actual advantages of face based identification over other biometrics are uniqueness and acceptance. As human face is a dynamic object having high degree of variability in its appearance, that makes face detection a difficult problem in computer vision. In this field, accuracy and speed of identification is a main issue.

The goal of this paper is to evaluate various face detection and recognition methods, provide complete solution for image based face detection and recognition with higher accuracy, better response rate as an initial step for video surveillance. Solution is proposed based on performed tests on various face rich databases in terms of subjects, pose, emotions, race and light.

## 1. Introduction

Over the last few decade lots of work is been done in face detection and recognition [14] as it's a best way for person identification [16] because it doesn't require human cooperation [15] so that it became a hot topic in biometrics. Since lots of methods are introduced for detection [6,7,8,12,13] and recognition [8,9,10,11] which considered as a milestone. Although these methods are used several times for the same purpose separately for limited number of datasets in past but there is no work found who provides overall performance evaluation of said methods altogether by testing them on tough datasets like [1,2,3,4,5], details of datasets will be provided in section IV.

In current paper we developed a system for the said method's evaluation as a first milestone for video based face detection and recognition for surveillance. The overview of current system is demonstrated in figure 1.

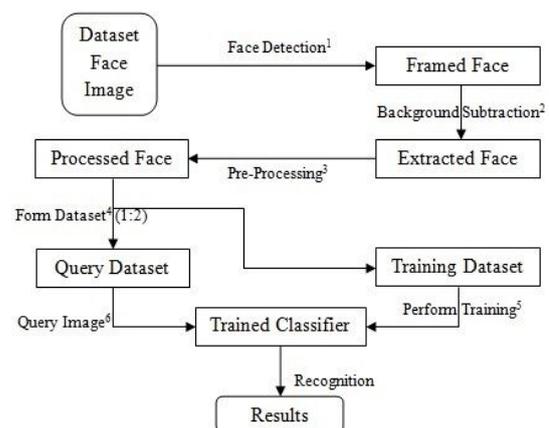

Fig. 1 System's overview.

1.1 Paper Organization

The following paper discuss about face detection methods in section II, section III discuss about face recognition methods based on the results of section II; result summery has been provided in the form of tables. Section IV provides summery about datasets that are been used in section II and section III. Section V is a conclusion phase.

## 2. Face Detection

AdaBoost [6] classifier is used with Haar [7] and Local Binary Pattern (LBP) [8] features whereas Support Vector Machine (SVM) [12] classifier is used with Histogram of Oriented Gradients (HOG) [13] features for face detection evaluation.

Haar-like [7] features are evaluated through the use of a new image representation that generates a large set of features and uses the boosting algorithm AdaBoost [6] to reduce degenerative tree of the boosted classifiers for robust and fast interferences only simple rectangular Haar-like [7] features are used that provides a number of benefits like sort of ad-hoc domain knowledge is implied as well as a speed increase over pixel based systems, suggestive to Haar [7] basis functions equivalent to intensity difference readings are quite easy to compute. Implementation of a system that used such features would provide a feature set that was far too large, hence the feature set must be only restricted to a small number of critical features which is achieved by boosting algorithm, Adaboost [6].

The original LBP [8] operator labels the pixels of an image by thresholding the 3-by-3 neighborhood of each pixel with the center pixel value and considering the result as a binary number. Each face image can be considered as a composition of micro-patterns which can be effectively detected by the LBP [8] operator. To consider the shape information of faces, they divided face images into N small non-overlapping regions $T_0, T_1, ..., T_N$. The LBP [8] histograms extracted from each sub-region are then concatenated into a single, spatially enhanced feature histogram defined as:

$$H_{i,j} = \Sigma_{x,y} I(f_l(x,y)=i) I((x,y) \epsilon T_j)$$

where $i = 0, ..., L-1$; $j = 0, ..., N-1$. The extracted feature histogram describes the local texture and global shape of face images.

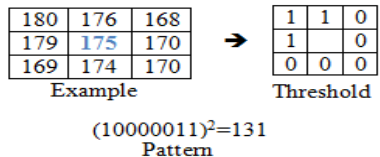

Fig. 2 LBP calculation.

SVM [12] classifier is been used with HOG [13] features for face detection. HOG [13] greatly outperforms wavelets and degree of smoothing before calculating gradients damages, results emphasizes much of the available information is from sudden edges at fine scales that blurring this for reducing the sensitivity to spatial position is a mistake. Gradients should be calculated at the finest available scale in the current pyramid layer and strong local contrast normalization is essential for good results. Whereas SVM [12] are formulated to solve a classical two class problem which returns a binary value, the class of the object. To train our SVM [12] algorithm, we formulate the problem in a difference space that explicitly captures the dissimilarity between two facial images. The results summery of above methods are stated below.

Table 1: Face detection results summery

| Dataset | Detection | | |
|---|---|---|---|
| | Adaboost | | SVM |
| | Haar | LBP | HOG |
| [1] | 99.31% | 95.22% | 92.68% |
| [2] | 98.33% | 98.96% | 94.10% |
| [3] | 98.31% | 69.83% | 87.89% |
| [4] | 96.94% | 94.16% | 90.58% |
| [5] | 90.65% | 88.31% | 89.19% |
| Mean | 96.70% | 89.30% | 90.88% |

Moreover system is been tested on datasets [1,2,3,4,5] and based on above demonstrated system, results are demonstrated below:

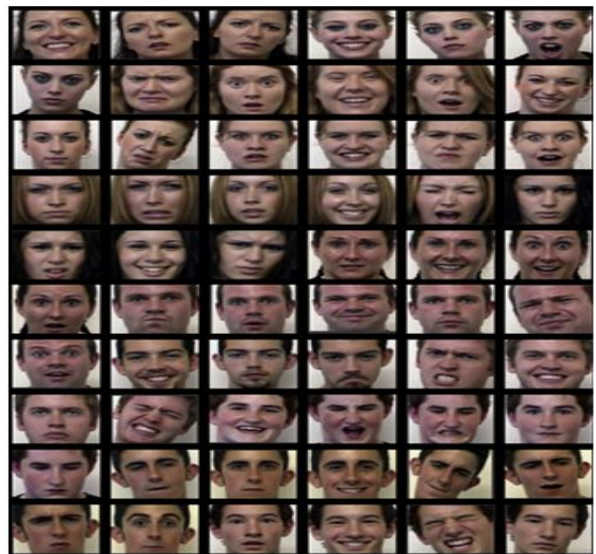

Fig. 3 Face detection.

To reduce pose variation and illumination in extracted faces two extra actions performed in pre-processing stage to improve recognition results: 1) Eyes detection is been used to remove head turn, tilt, slant and position of face, demonstrated in figure 4; 2) Histogram equalization is been performed.

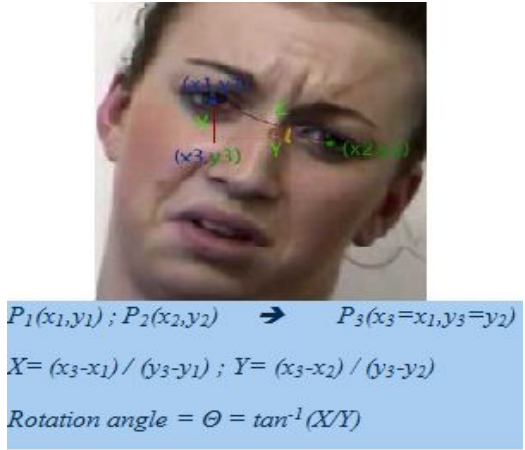

$P_1(x_1,y_1) ; P_2(x_2,y_2) \rightarrow P_3(x_3=x_1, y_3=y_2)$

$X= (x_3-x_1) / (y_3-y_1) ; Y= (x_3-x_2) / (y_3-y_2)$

Rotation angle = $\Theta = \tan^{-1}(X/Y)$

**Fig. 4** Face rotation.

## 3. Face Recognition

Eigenfaces [9] considered as 2-D face recognition problem, faces will be mostly upright and frontal. That's why 3-D information about the face is not required that reduces complexity by a significant bit. It convert the face images into a set of basis functions which essentially are the principal components of the face images seeks directions in which it is more efficient to represent the data. This is mainly useful for decrease the computational effort. Linear discriminant analysis is primarily used here to reduce the number of features to a more manageable number before recognition because face is represented by a large number of pixel values. Each of the new dimensions is a linear combination of pixel values, which form a template. The linear combinations obtained using Fisher's linear discriminant are called Fisherfaces [10]. LBP [8] is an order set of binary comparisons of pixel intensities between the center pixel and its eight surrounding pixels.

$$LBP (x_a, y_a) = {}^7\Sigma_{n=0}\, s(i_m - i_a)\, 2^n$$

Where $i_a$ corresponds to the value of the center pixel $(x_a, y_a)$, $i_m$ to the value of eight surrounding pixels, function $f(x)$ is defined as:

$$f(x) = \begin{cases} 1 & \text{if } x >= 0 \\ 0 & \text{if } x < 0 \end{cases}$$

Gabor [11] filters can exploit salient visual properties such as spatial localization, orientation selectivity, and spatial frequency characteristics. Considering these devastating capacities and its great success in face recognition Gabor [11] features are insensitive to transformations as illumination, pose and expressions although Gabor [11] transform is not specially designed for face recognition. Its transformation formula is predefined instead of learned from the face training data. Moreover PCA [9] and LDA [10] classifier consider global features whereas LBP [8] and Gagor classifier consider local features, based on current facts experimental results are stated below.

Table 2: Face recognition results summery

| Dataset | Recognition | | | |
|---|---|---|---|---|
| | PCA | LDA | LBP | Gabor |
| [1] | 72.10% | 79.39% | 85.93% | 93.49% |
| [2] | 69.87% | 76.61% | 80.47% | 89.76% |
| [3] | 70.95% | 78.34% | 84.14% | 92.68% |
| [4] | 74.79% | 81.93% | 86.45% | 96.91% |
| [5] | 68.04% | 73.21% | 77.69% | 88.93% |
| Mean | 71.15% | 77.90% | 82.94% | 92.35% |

## 4. Dataset

Five datasets been used for above experiments. In dataset [1], face collection with plain green background; no head scale and light variation but having minor changes in head turn, tilt, slant, position of face and considerable change in expressions.

In dataset [2], face collection with red curtain background, variation is caused by shadows as subject moves forward, having minor changes in head turn, tilt and slant; large head scale variation; some expression variation, translation in position of face and image lighting variation as subject moves forward, significant lighting changes occur on faces moment due to the artificial lighting arrangement. In dataset [3], face collection with complex background; large head scale variation; minor variations in head turn, tilt, slant and expression; some translation in face position and significant light variation because of object moment in artificial light. In dataset [4], face collection with plain background; small head scale variation; considerable variation in head turn, tilt, slant and major variation in expression; minor translation in face position and light variation. In dataset [5], face collection with constant background having minor head scale variation and light variation; huge variation in turn, tilt, slant, expression and face position.

## 5. Conclusion

In current work we developed the system to evaluate the face detection and recognition methods which are

Table 3: Face database summery

| Data Set | Sub-Division | Images | Resolution | Individuals | Image/Individual |
|---|---|---|---|---|---|
| A | Face 94 | 3078 | 180*200 | 153 | ~20 |
| A | Face 95 | 1440 | 180*200 | 72 | 20 |
| A | Face 96 | 3016 | 196*196 | 152 | ~20 |
| A | Grimace | 360 | 180*200 | 18 | 20 |
| B | Pain Expressions | 599 | 720*576 | 23 | 26 |

A: Face Recognition Data, University of Essex
B: Psychological Image Collection at Stirling (PICS)

considered to be a bench mark. Some methods performed consistently over different datasets whereas other methods behave very randomly however based on average experimental results performance is evaluated, five datasets been used for this purpose. Face detection and recognition method's result summery is provided in table 1 and table 2 respectively whereas datasets summery is provided in table 3. In current system Haar-like [7] features reported relatively well but it has much false detection than LBP [8] which could be consider being a future work in surveillance to reduce false detection in Haar-like [7] features and for the recognition part gabor [11] is reported well as it's qualities overcomes datasets complexity.

**Acknowledgment**

The work is being mainly done at Laboratory of Intelligent Recognition and Image Processing, Beijing University of Aeronautics and Astronautics. We would like to thank Dr Zhaoxiang Zhang for his guidance and support. He encouraged us to write this research paper. We are also grateful to our family for putting up with us.

## References

[1] Face Recognition Data, University of Essex, UK, Face 94, http://cswww.essex.ac.uk/mv/all faces/faces94.html.

[2] Face Recognition Data, University of Essex, UK, Face 95, http://cswww.essex.ac.uk/mv/all faces/faces95.html.

[3] Face Recognition Data, University of Essex, UK, Face 96, http://cswww.essex.ac.uk/mv/all faces/faces96.html.

[4] Face Recognition Data, University of Essex, UK, Grimace, http://cswww.essex.ac.uk/mv/all faces/grimace.html.

[5] Psychological Image Collection at Stirling (PICS), Pain Expressions, http://pics.psych.stir.ac. uk/2D_face_sets.htm.

[6] K. T. Talele, S. Kadam, A. Tikare, Efficient Face Detection using Adaboost, "IJCA Proc on International Conference in Computational Intelligence", 2012.

[7] T. Mita, T. Kaneko, O. Hori, Joint Haar-like Features for Face Detection, "Proceedings of the Tenth IEEE International Conference on Computer Vision", 1550-5499/05 ©2005 IEEE.

[8] T. Ahonen, A. Hadid, M. Peitikainen, Face recognition with local binary patterns. "In Proc. of European Conference of Computer Vision", 2004.

[9] M. A. Turk and A.P. Pentland, Face recognition using eigenfaces, "Proceedings of the IEEE", 586-591, 1991.

[10] J Lu, K. N. Plataniotis, A. N. Venetsanopoulos, Face recognition using LDA-based algorithms, "IEEE Neural Networks Transaction", 2003.

[11] L. Wiskott, M. Fellous, N. Krger, and C. Malsburg, Face recognition by elastic bunch graph matching, "IEEE Trans", on PAMI, 19:775–779, 1997.

[12] I. Kukenys, B. McCane, Support Vector Machines for Human Face Detection, "Proceedings of the New Zealand Computer Science Research Student Conference", 2008.

[13] M. M. Abdelwahab, S. A. Aly, I. Yousry, Efficient Web-Based Facial Recognition System Employing 2DHOG, arXiv:1202.2449v1 [cs.CV].

[14] W. Zhao, R. chellappa, P. J. Phillips, Face recognition: A literature survey, "ACM Computing Surveys (CSUR)", December 2003.

[15] G. L. Marcialis, F. Roli, Chapter: Fusion of Face Recognition Algorithms for Video-Based Surveillance Systems, Department of Electrical and Electronic Engineering- Univ- ersity of Cagliari- Italy.

[16] A. Suman, Automated face recognition: Applications within law enforcement. Market and technology review, "NPIA", 2006.